%% file: iccp24_template.tex
\documentclass[10pt,journal,compsoc]{IEEEtran}
\newif\ifpeerreview

\peerreviewfalse

\usepackage[nocompress]{cite}
\usepackage{url}
\usepackage{amsmath,amssymb,graphicx}

\usepackage{lipsum} 
\usepackage{cleveref}
\usepackage[numbers,sort&compress]{natbib}

\usepackage[dvipsnames]{xcolor}

\usepackage[switch]{lineno}

\usepackage{algorithm}
\usepackage{algpseudocode}
\usepackage{booktabs} 
\usepackage{tabularx}
\usepackage{threeparttable}

\usepackage{algorithm}
\usepackage{algpseudocode}

\usepackage{ulem}

\newcommand{\qq}[1]{{\color{black} #1}}
\newcommand{\gs}{\text{gs}}
\definecolor{mydarkgreen}{rgb}{0.0, 0.5, 0.0}
\newcommand{\paperID}{41}

\title{Z-Splat: Z-Axis Gaussian Splatting for Camera-Sonar Fusion}

\author{\IEEEauthorblockN{
Ziyuan Qu\IEEEauthorrefmark{1},
Omkar Vengurlekar\IEEEauthorrefmark{2}, 
Mohamad Qadri\IEEEauthorrefmark{3}, 
Kevin Zhang\IEEEauthorrefmark{4},
Michael Kaess\IEEEauthorrefmark{3},
Christopher Metzler\IEEEauthorrefmark{4},
Suren Jayasuriya\IEEEauthorrefmark{2}, and
Adithya Pediredla\IEEEauthorrefmark{1} \\
}
\IEEEauthorblockA{\IEEEauthorrefmark{1}Dartmouth College,}
\IEEEauthorblockA{\IEEEauthorrefmark{2}Arizona State University,}
\IEEEauthorblockA{\IEEEauthorrefmark{3}Carnegie Mellon University,}
\IEEEauthorblockA{\IEEEauthorrefmark{4}University of Maryland}}

\begin{document}    

\input{sections/00_abstract}

\maketitle

\input{sections/01_introduction}

\input{sections/02_related_work}
\input{sections/03_background}

\input{sections/04_proposed_method}

\input{sections/05_experimental_results}


\input{sections/07_conclusion}

\ifpeerreview \else
\section*{Acknowledgments}
A.P. and Z.Q. were supported in part by NSF CCF-2326904.
O.V. and S.J. were supported in part by ONR grant N00014-23-1-2406, NSF CCF-2326905, and Raytheon, Inc. In addition, the authors acknowledge Research Computing at Arizona State University for providing the Sol supercomputer~\cite{HPC:ASU23} for GPU computing. M.Q. and M.K. were supported in part by ONR grant N00014-21-1-2482.
K.Z.~and C.A.M.~were supported in part by AFOSR Young Investigator Program award no. FA9550-22-1-0208, ONR award no. N00014-23-1-2752, and a seed grant from SAAB, Inc. 
\fi
\small
\bibliographystyle{IEEEtran}
\bibliography{references}

\ifpeerreview \else


\vspace{-4\baselineskip}
\begin{IEEEbiography}[{\includegraphics[width=1in,height=1.25in,clip,keepaspectratio]{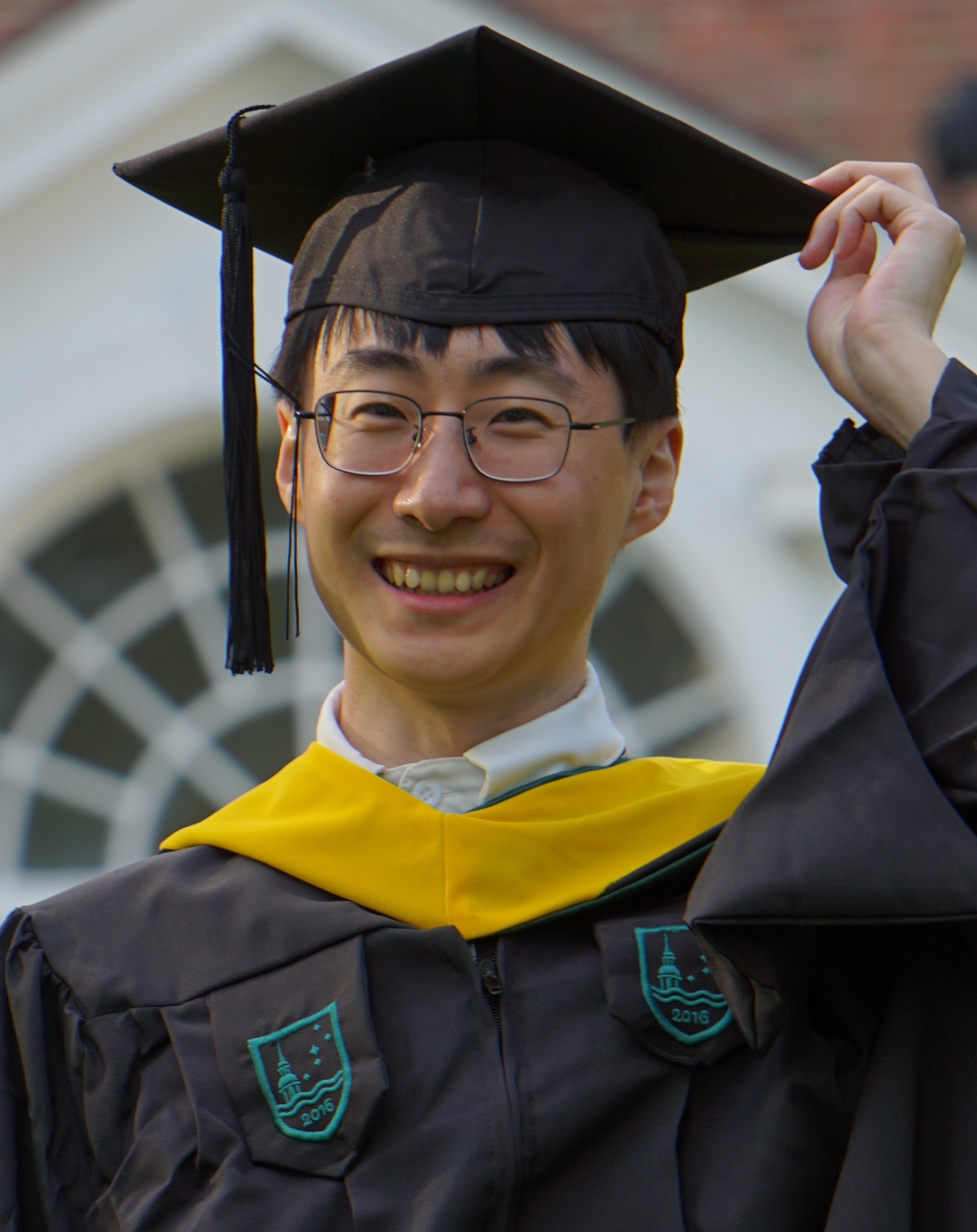}}]{Ziyuan Qu}
received his B.Eng. in Software Engineering from Northeastern University, China in 2022, and his M.S. in Computer Science from Dartmouth College in 2024. He is currently pursuing a PhD in Computer Science at Dartmouth College under supervision of Prof. Adithya Pediredla, focusing on computational photography and rendering.
\end{IEEEbiography}
\vspace{-4\baselineskip}
\begin{IEEEbiography}[{\includegraphics[angle=-89.999,width=1in,height=1.25in,clip,keepaspectratio]{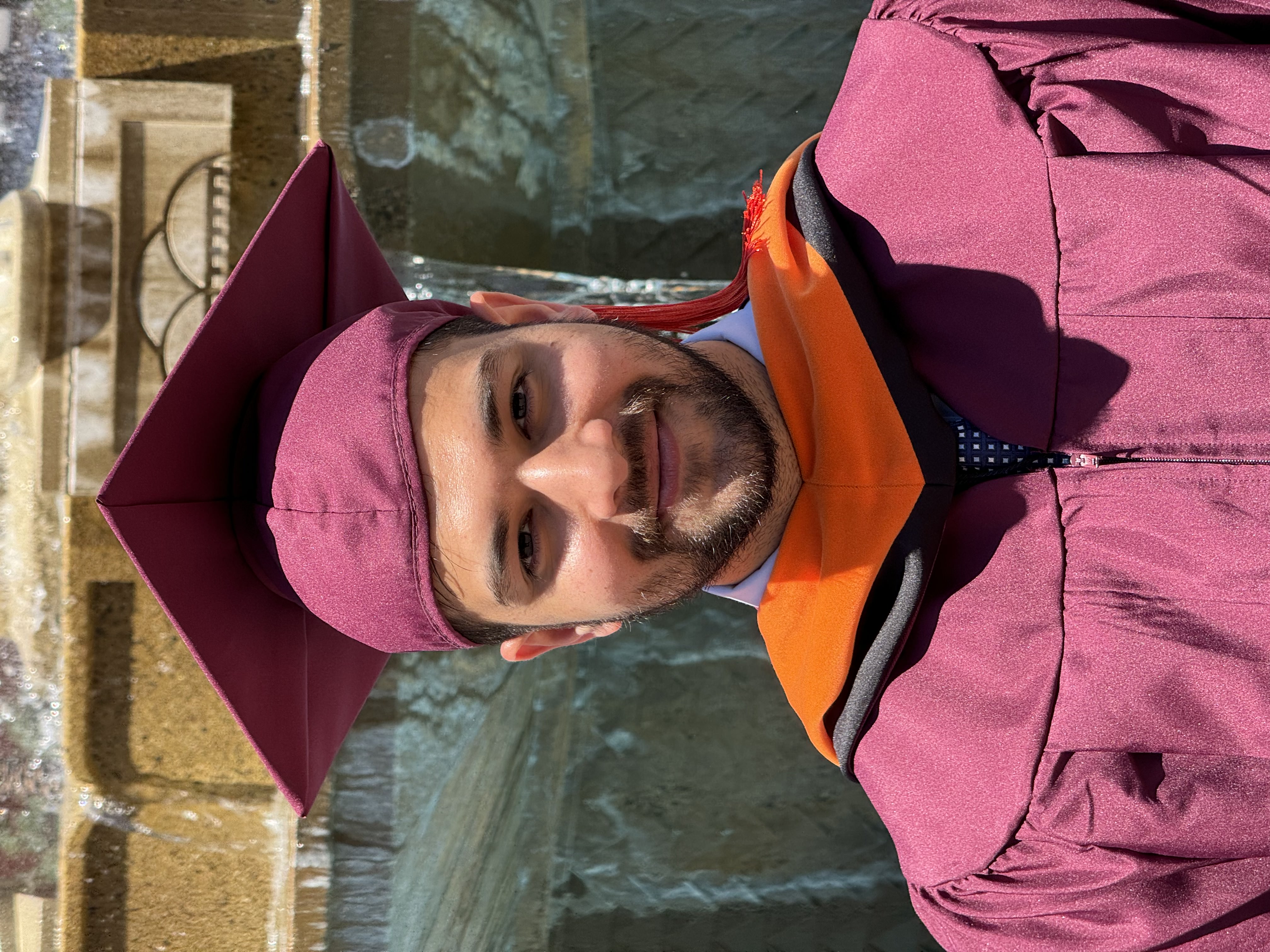}}]{Omkar Vengurlekar}
received his B.Eng. in Electronics and Telecommunication Engineering from Pune University, India in 2020, and his M.S. in Robotics and Autonomous Systems from Arizona State University in 2024. He is currently pursuing a PhD in Computer Engineering at Arizona State University under the supervision of Prof. Suren Jayasuriya, focusing on computational vision, 3D Reconstruction, and machine learning.
\end{IEEEbiography}
\vspace{-4\baselineskip}
\begin{IEEEbiography}[{\includegraphics[width=1in,height=1.25in,clip,keepaspectratio]{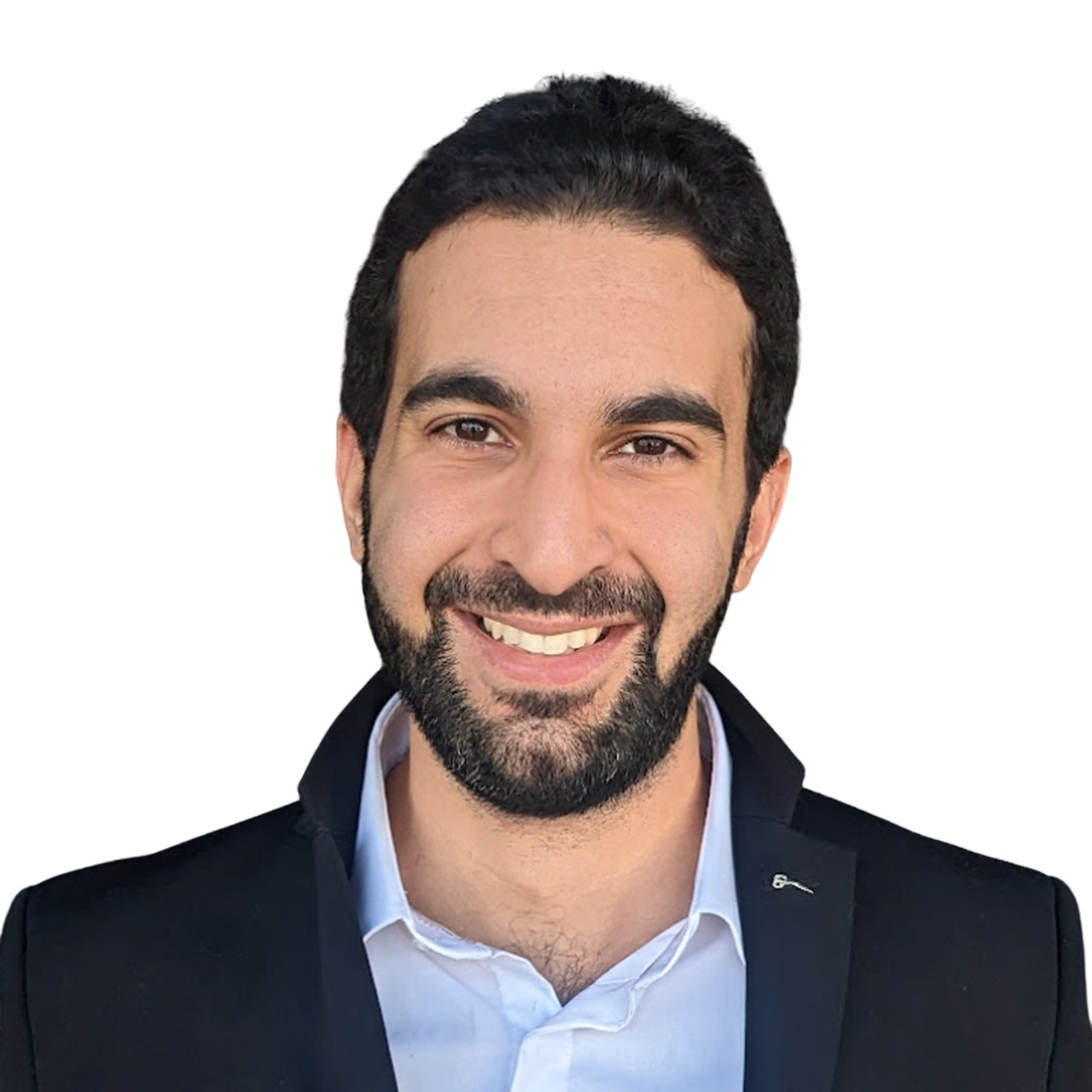}}]{Mohamad Qadri}
received his BS in Electrical Engineering from the University of Maryland, College Park in 2016, and his MS in Robotics from Carnegie Mellon University in 2021. He is currently pursuing a PhD in Robotics at Carnegie Mellon University with a focus on computer vision. 
\end{IEEEbiography}
\vspace{-3\baselineskip}
\begin{IEEEbiography}[{\includegraphics[width=1in,height=1.25in,clip,keepaspectratio]{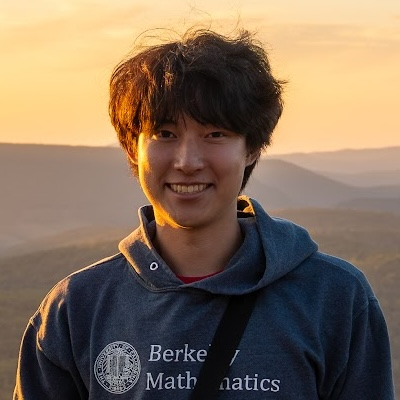}}]{Kevin Zhang}
received the B.A. in Computer Science and Pure Mathematics from the University of California, Berkeley. He is currently a Computer Science Ph.D. student at the University of Maryland, College Park, based primarily in the Intelligent Sensing Lab, advised by Professor Christopher Metzler and Jia-bin Huang. His research focuses on 3D reconstruction from various sensing modalities.
\end{IEEEbiography}
\vspace{-3\baselineskip}
\begin{IEEEbiography}[{\includegraphics[width=1in,height=1.25in,clip,keepaspectratio]{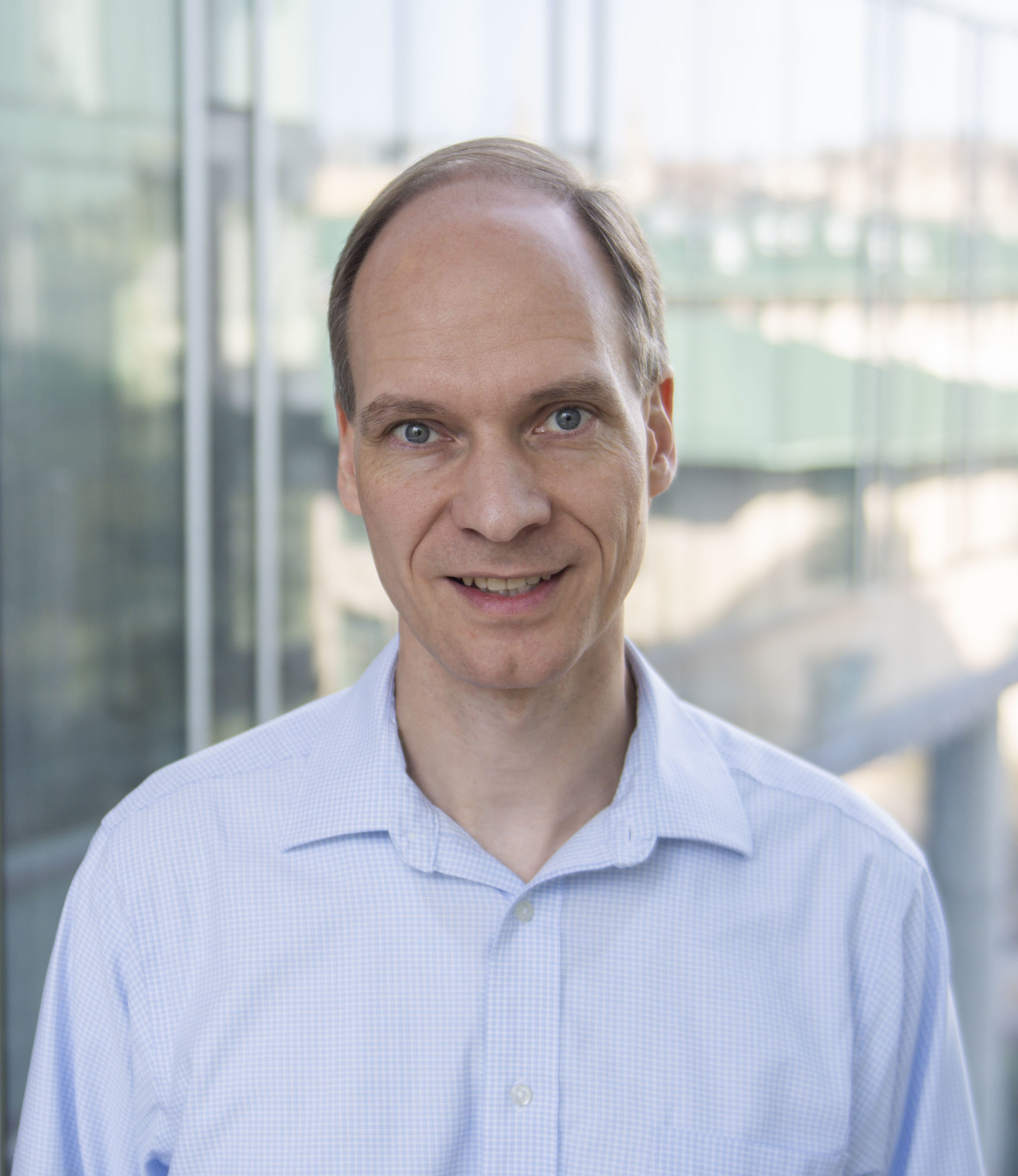}}]{Michael Kaess}
(Senior Member, IEEE) received the M.S. and Ph.D. degrees in computer science from the Georgia Institute of Technology, Atlanta, GA, USA, in 2002 and 2008, respectively.
He is an Associate Professor in the Robotics Institute at Carnegie Mellon University, Pittsburgh, PA, USA. Before he was a postdoctoral associate and research scientist at the Computer Science and Artificial Intelligence Laboratory (CSAIL) at the Massachusetts Institute of Technology (MIT), Cambridge, MA, USA. He is currently the Directory of the Robot Perception Lab, Carnegie Mellon University. His research interests include state estimation, localization, mapping, navigation, and autonomy.
Dr. Kaess is a member of IEEE, AAAI, and ACM. He is the recipient of the inaugural Robotics: Science and Systems Test of Time Award 2020.
\end{IEEEbiography}
\vspace{-2.5\baselineskip}
\begin{IEEEbiography}[{\includegraphics[width=1in,height=1.25in,clip,keepaspectratio]{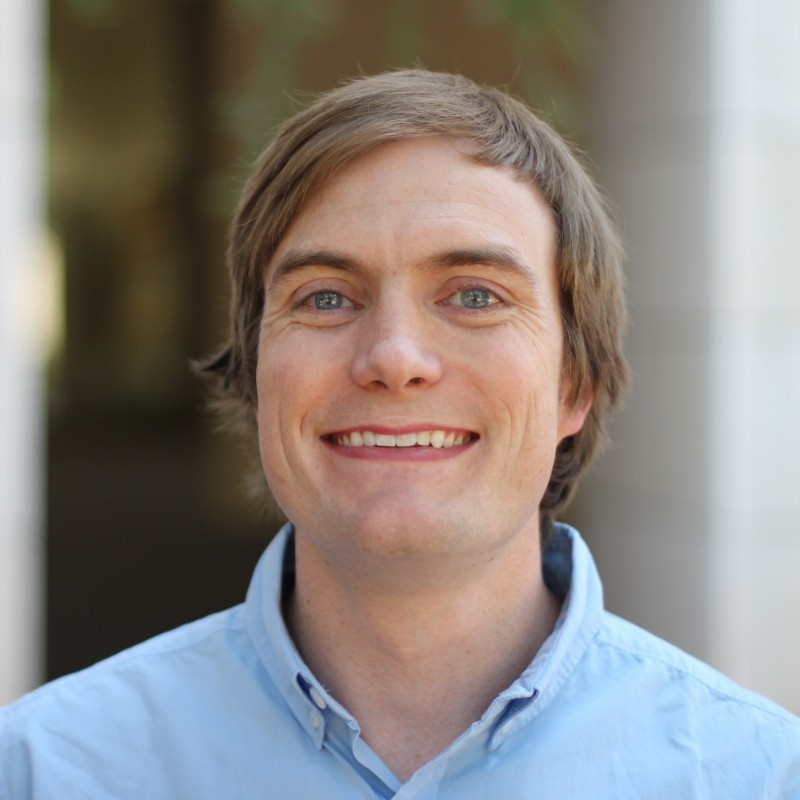}}]{Christopher A. Metzler}
is an Assistant Professor in the Department of Computer Science at the University of Maryland College Park, where he leads the UMD Intelligent Sensing Laboratory. He is a member of the University of Maryland Institute for Advanced Computer Studies (UMIACS) and has a courtesy appointment in the Electrical and Computer Engineering Department. His research develops new systems and algorithms for solving problems in computational imaging and sensing, machine learning, and wireless communications. His work has received multiple best paper awards; he recently received NSF CAREER, AFOSR Young Investigator Program, and ARO Early Career Program awards; and he was an Intelligence Community Postdoctoral Research Fellow, an NSF Graduate Research Fellow, a DoD NDSEG Fellow, and a NASA Texas Space Grant Consortium Fellow.
\end{IEEEbiography}
\vspace{-2.5\baselineskip}
\begin{IEEEbiography}[{\includegraphics[width=1in,height=1.25in,clip,keepaspectratio]{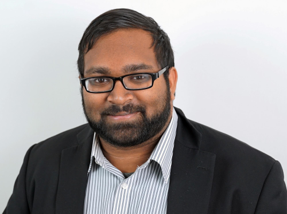}}]{Suren Jayasuriya}
(Senior Member, IEEE) is an assistant professor at Arizona State University, in the School of Arts, Media and Engineering (AME) and Electrical, Computer and Energy Engineering (ECEE) since 2018. Before this, he was a postdoctoral fellow at the Robotics Institute at Carnegie Mellon University from 2016-2017. He received his Ph.D. in electrical and computer engineering at Cornell University in 2017 and graduated from the University of Pittsburgh in 2012 with a B.S. in Mathematics (with departmental honors) and a B.A. in Philosophy. His research interests range from computational imaging and photography, computer vision and graphics, and machine learning. 
\end{IEEEbiography}
\vspace{-2.5\baselineskip}
\begin{IEEEbiography}[{\includegraphics[width=1in,height=1.25in,clip,keepaspectratio]{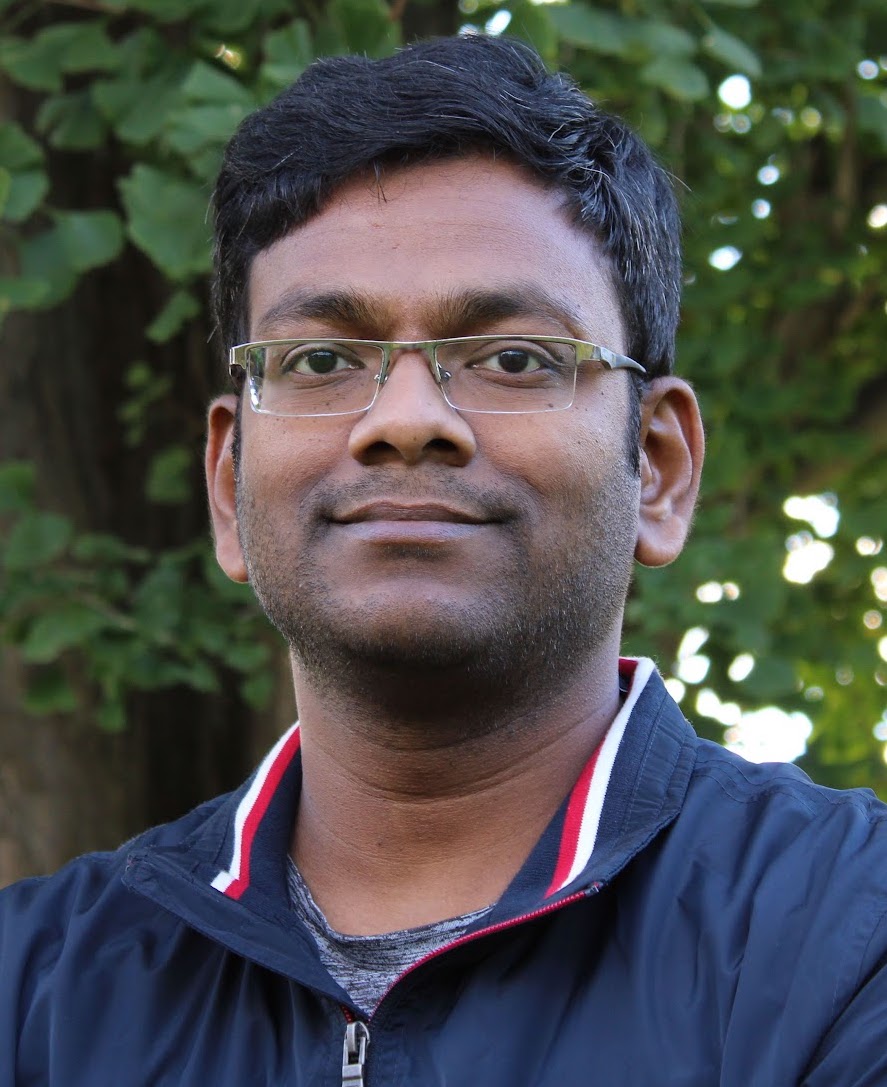}}]{Adithya Pediredla}
is an Assistant Professor in the Department of Computer Science at Dartmouth College, where he leads the Rendering and Imaging Science (RISC) lab, focusing on the intersection of computer graphics, computational imaging, and computer vision. Before this role, he is a postdoctoral fellow at the Robotics Institute, Carnegie Mellon University. He has a Ph.D. from Rice University and a master's degree from the Indian Institute of Science. He won the Ralph Budd Best Engineering Thesis Award for his Ph.D. thesis, the Prof. K. R. Kambati Memorial gold medal, and an innovative student project award from the Indian National Academy of Engineering for his Master's thesis. 
\end{IEEEbiography}
\vfill




\fi

\end{document}

%% file: sections/00_abstract.tex
\IEEEtitleabstractindextext{%
\begin{abstract}
Differentiable 3D-Gaussian splatting (GS) is emerging as a prominent technique in computer vision and graphics for reconstructing 3D scenes. GS represents a scene as a set of 3D Gaussians with varying opacities and employs a computationally efficient splatting operation along with analytical derivatives to compute the 3D Gaussian parameters given scene images captured from various viewpoints. Unfortunately, capturing surround view ($360^{\circ}$ viewpoint) images is impossible or impractical in many real-world imaging scenarios, including underwater imaging, rooms inside a building, and autonomous navigation. In these restricted baseline imaging scenarios, the GS algorithm suffers from a well-known `missing cone' problem, which results in poor reconstruction along the depth axis. In this paper, we demonstrate that using transient data (from sonars) allows us to address the missing cone problem by sampling high-frequency data along the depth axis. We extend the Gaussian splatting algorithms for two commonly used sonars and propose fusion algorithms that simultaneously utilize RGB camera data and sonar data. Through simulations, emulations, and hardware experiments across various imaging scenarios, we show that the proposed fusion algorithms lead to significantly better novel view synthesis ($5$~dB improvement in PSNR) and 3D geometry reconstruction (60\% lower Chamfer distance).

\end{abstract}

\begin{IEEEkeywords} 
Gaussian Splatting, Acoustic-Optic Vision, Sensor fusion.
\end{IEEEkeywords}
}

%% file: sections/01_introduction.tex
\IEEEraisesectionheading{
  \section{Introduction}\label{sec:introduction}
}
%
%
%
%
\IEEEPARstart{D}{ifferentiable} Gaussian spatting (GS)~\cite{kerbl20233d}, a resurgence of elliptical weighted average (EWA) splatting~\cite{zwicker2001ewa} from almost two decades back, has been emerging as the dominant differentiable representation for scenes~\cite{lu2023scaffold, yu2023mip}. 
GS represents the scene as a volume density map similar to neural radiance fields (Nerf)~\cite{kerbl20233d, mildenhall2021nerf}. 
However, unlike Nerfs, GS explicitly represents the scene as a sum of densities of anisotropic 3D Gaussians. 
The explicit representation facilitates easier geometric interpretation and manipulation of the scene, leading to an explosion of GS-based techniques for scene relighting~\cite{gao2023relightable}, shading~\cite{jiang2023gaussianshader}, novel view synthesis~\cite{zhu2023fsgs}, text-based manipulation~\cite{fang2023gaussianeditor}, and non-rigid manipulation~\cite{das2023neural}. 
Furthermore, the GS rendering algorithm is fast, thanks to the splatting operator and analytical derivatives which leads to extensions to dynamic scenes~\cite{yang2023real, kratimenos2023dynmf}, 4D manipulation~\cite{shao2023control4d, huang2023sc, yu2023cogs} and time-efficient editing~\cite{huang2023point,fang2023gaussianeditor}.  


Unfortunately, when the training dataset has a limited baseline, optimizing through the GS algorithm causes overfitting leading to poor novel-view rendering and 3D reconstruction. 
In \Cref{sec:missing_cone}, we show that limited baseline results in the well-known `missing cone' problem \cite{macias1988missing,liu2019analysis} where the frequencies along the depth-axis are not captured in the training data. 
To mitigate the problem of missing depth information from the training images, Chung et al.~\cite{chung2024depth} proposed regularizing 3D Gaussian splatting with monocular depth estimates.
However, as this method hallucinates depth, the scene reconstructions are no longer physically based and suffer from training bias. In this paper, we use sonar images to sample the missing cone region. 

Recent progress in sonar technologies has resulted in several low-cost sensors that augment 2D spatial data measured by RGB cameras. 
Examples include echosounder~\cite{liu2023deep}, forward-looking sonar~\cite{qadri2024aoneus}, synthetic aperture sonar~\cite{reed2023neural} which find applications in SLAM~\cite{fallon2013relocating}, navigation~\cite{yang2022monocular, lin2023conditional}, underwater imaging~\cite{roznere2020underwater}, and imaging through scattering media~\cite{zhang2019underwater}. 
These sonars offer complementary information compared to the standard RGB cameras. 

In this paper, we extend Gaussian splatting for sonar and build fusion techniques that reconstruct geometry using the complementary information from both the cameras and sonars. 
Our extension involves the development of splatting operations along the $z$-axis tailored for these sensor types.
Through simulations, emulations, and hardware experiments, we show that combining sonars and cameras results in significantly better (60\%) 3D geometry reconstruction. We also show that the accuracy of novel view synthesis improves by 5~dB by regularizing camera data with sonars.


The specific contributions of this paper include
\begin{enumerate}
    \item A novel forward model to render the transient of Gaussian point clouds for two types of sonars: Echosounder and Forward-Looking Sonar (FLS).
    \item Fusion algorithms for cameras and sonars. 
    \item Validation on synthetic, emulated hardware, and real hardware datasets showing that fusion GS splatting results in better geometric (60\%) and photometric (5 dB) reconstruction than standard camera-only Gaussian splatting. 
\end{enumerate}

\qq{We release our source code and data to the larger community \cite{zsplatGitHub}. We hope that our work inspires the use of complementary sensor modalities to improve 3D scene representation in the future.}




%% file: sections/02_related_work.tex
\section{Related Work}

{\bf Gaussian splatting:}
In computer graphics and rendering, splatting algorithms were introduced over two decades ago~\cite{zwicker2001ewa} for texture filtering~\cite{heckbert1989fundamentals} and point cloud rendering~\cite{zwicker2001surface, zwicker2002ewa}, where the scene is represented as a sum of anisotropic Gaussian kernel that can be efficiently rendered and without aliasing artifacts. 

The Gaussian splatting paper~\cite{kerbl20233d} uses this scene representation and fast differentiable rendering pipeline to compute the scene parameters (density, color, mean, and variance of the 3D Gaussians). Thanks to the decreased computation on empty spaces, analytical derivatives, rich-quality reconstructions, and explicit representations, Gaussian splatting has seen an explosion of extensions, even though it was introduced less than a year ago (see Fei et al.~\cite{fei20243d} for a review of the state-of-the-art). 
Related to this paper, Matsuki et al.~\cite{matsuki2023gaussian},  Keetha et al.~\cite{keetha2023splatam}, Yan et al.~\cite{yan2023gs}, Sun et al.~\cite{sun2024high} have independently proposed similar SLAM algorithms using RGB-D cameras. 
The key idea is to splat depth similar to color values and minimize the weighted sum of photometric and geometric loss functions. However, this idea cannot be extended to sonar data as these techniques require a single depth value per pixel, necessitating the dataset to be a depth map with high spatial resolution on the x-y plane. Instead, sonar data typically resembles a transient histogram of time-of-flight returns. 


\noindent {\bf Camera and sonar fusion techniques:}
There has been previous research on fusing complementary information from sonars and cameras in the literature~\cite{Ferreira2016, Raaj2016,williams2004simultaneous}. For example, optical cameras suffer from hazing and scattering problems, especially in turbid waters~\cite{Ferreira2016}. Sonars do not suffer from scattering, though they have poor spatial resolution. 
To address this, Raaj et al.~\cite{Raaj2016} combined FLS and optical cameras for 3D object localization via particle filter in scattering environments. Williams and Mahon~\cite{williams2004simultaneous} combined an echosounder (also known as depth sounder) and a vision camera to develop a SLAM algorithm for underwater robotic navigation on the Great Barrier Reef. This paper focuses on the complementary geometric information that sonars and cameras provide and analyzes how the missing cone present in camera-based 3D reconstruction can be resolved via our Gaussian splatting algorithms for transient histograms to reconstruct better geometry and photometry. 


Most related to our work, Babaee and Negahdaripour~\cite{babaee20153} reconstructed 3D objects from RGB and imaging sonars. However, their technique requires matching occluding contours across RGB and imaging sonar, which restricts its applicability to small baseline scenarios. Qadri et al.~\cite{qadri2024aoneus} recently propose combining FLS sonar and cameras using implicit neural representations (INR). Unlike ours, Qadri et al.'s technique cannot easily extend to echosounder and reconstructs only scene geometry, not photometry. 

%

\noindent{\bf Other fusion techniques:} 
In addition to sonar, researchers have proposed fusing radar measurements with radar for better object detection~\cite{nabati2021centerfusion}, imaging of cluttered environment~\cite{grover2001low}, and estimating pose and motion of vehicles for autonomous navigation. Fusing Lidars or other optical time-of-flight cameras is similar to our problem and has been targeted at imaging through scattering media~\cite{bijelic2020seeing}, densification of lidar point scans~\cite{kim2009multi}, and improved depth estimation~\cite{lindell2018single, nishimura2020disambiguating, attal2021torf}. The proposed Gaussian splatting algorithm could be extended to fusion with Radars and Lidars. 

%% file: sections/03_background.tex
\section{Technical background}
\label{sec:technical_background}

\begin{figure}[t]
  \centering
  \setlength{\abovecaptionskip}{-0.12cm}
   \includegraphics[width=\linewidth]{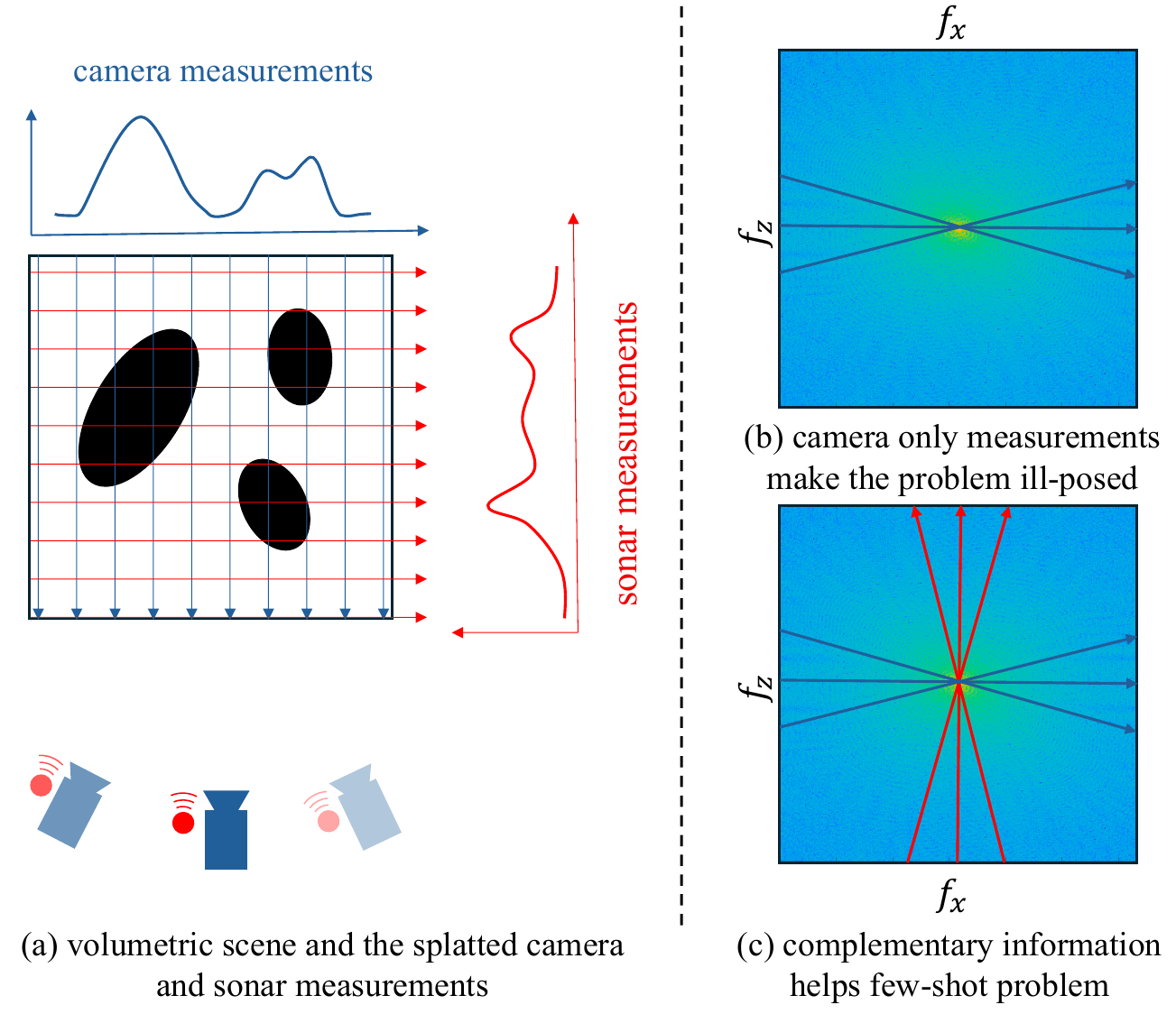}
   \caption{\textbf{Sonar measurements provide complementary information.} (a) Volumetric scene captured with three pairs of cameras and sonars (echosounder). We assume the sensors are in the far field (i.e., the affine approximation to the projective transform in Gaussian splatting research is valid). For the center camera-sonar pair, camera measurements are obtained by projecting the volumetric data along the vertical axis, and sonar measurements are obtained by projecting the volumetric data along the horizontal axis. (b) If only camera measurements are considered, then using the Fourier-slice theorem, we are capturing only a few slices of the Fourier transform of the volume and missing information on a large cone. (c) Sonar (time-resolved data) captures orthogonal slices in the Fourier space, and hence,  3D reconstruction of the scene is better conditioned if we do the camera-sensor fusion instead of using only camera data. 
   }
   \label{fig:fusion_rationale}
\end{figure}
In this section, we will briefly review the scene representation, rendering equation, and splatting operation associated with the Gaussian splatting algorithm. 
We will show that traditional GS algorithms do not appropriately capture the $z$-related scene parameters, resulting in the missing cone problem. This problem motivates our camera and sonar fusion solution to recover this lost information.

\subsection{Scene representation}

In Gaussian splatting, the scene is represented as a volumetric density map. This density ($\sigma \in \mathbb{R}^+$) at a point ($\bar{x} \in \mathbb{R}^3$) is given as the sum of densities contributed by several Gaussians located as positions ($\bar{\mu}_n$ $\in \mathbb{R}^3$) and $3 \times 3$ anisotropic variance ($\varSigma_n$). Analytically, 
\begin{align}
    \sigma(\bar{x}) = \sum_{n=1}^{N} 
    \sigma_n  \mathcal{N}(\bar{x}; \bar{\mu}_n, \varSigma_n),
    \label{GS}
\end{align}
where $\sigma_n$ is a scaling factor representing the density of each Gaussian. The color at each 3D point is similarly defined as 
\begin{align}
    \bar{c}(\bar{x}) = \sum_{n=1}^{N} 
    \bar{c}_n  \mathcal{N}(\bar{x}; \bar{\mu}_n, \varSigma_n),
    \label{GS_color}
\end{align}
where $\bar{c}_n$ is the color of each Gaussian kernel, which is encoded using spherical harmonics \cite{kerbl20233d}.

The covariance of a 3D Gaussian is represented using the scaling matrix $S_n$ and a rotation matrix $R_n$ to maintain the positive definiteness as:
\begin{equation}
  \varSigma_n = R_n S_n S_n^T R_n^T.
  \label{eq:3D_Gaussian_cov_decompose}
\end{equation}
The rotation matrix $R_n$ is calculated from a normalized quaternion $q_n$. For brevity, we will drop the index $n$ wherever the index is implied. 

\subsection{Volume rendering equation}

Gaussian splatting uses the same volume rendering equation as point-based $\alpha$-blending~\cite{salvi2014multi} or NeRF-style~\cite{mildenhall2021nerf} volumetric rendering. Therefore, if a ray ($\bar{x}(l) = \bar{o} + l\bar{d} $) is cast along a pixel on the camera, the color $C$ of the pixel will be
\begin{align}
 C = \sum_{m=1}^M T_m \alpha_m c_m, 
 \label{eq:alpha_blending}
\end{align}
where the index $m$ represents the $m^{\text{th}}$ segment of the ray with small length $\delta l$, and
\begin{align}
\alpha_m &= 1-e^{-\sigma(\bar{x})\delta l},\, \,\, T_m = \prod_{k=1}^{m-1}(1-\alpha_k), \text{and} \nonumber \\ 
c_m &= c(\bar{x}(m\delta_l)).
\end{align}


\subsection{Splatting operation}
Instead of expensive ray tracing similar to NeRF-style algorithms, the GS algorithms~\cite{zwicker2001ewa, kerbl20233d} use splatting, which is similar to rasterization. 
For a camera viewing transform $W$, the 3D volume is transformed appropriately to the camera view, but the GS algorithm approximates the projection operation with an affine transformation so that any 3D Gaussian remains a 3D Gaussian with covariance matrix: 
\begin{align}
    \varSigma' = J W \varSigma W^T J^T,\,\text{and mean}\,\mu' = J W \mu
\end{align}
where $J$ is the Jacobian for the \qq{local affine approximation for each Gaussian kernel}
\begin{equation}
    J = 
    \begin{bmatrix}
         \frac{1}{\mu_z} & 0 & \frac{-\mu_x}{\mu_z^2}  \\
         0 & \frac{1}{\mu_z} & \frac{-\mu_y}{\mu_z^2}  \\
         \frac{\mu_x}{l} & \frac{\mu_y}{l} & \frac{\mu_z}{l},
    \end{bmatrix}
    \label{eq:Jacobian}
\end{equation}
and where $\mu = [\mu_x, \mu_y, \mu_z]$ and $l= \|\mu\|_2$.
\qq{The local affine approximation preserves the Euclidean distance from the camera to the object. Based on the equation (15) in the EWA Splatting \cite{zwicker2001ewa}, the transformed domain is 
\begin{align}
    \begin{bmatrix}
        \mu'_x &
        \mu'_y &
        \mu'_z 
    \end{bmatrix}^T
    =
    \begin{bmatrix}
        \mu_x/\mu_z &
        \mu_y/\mu_z &
        ||(\mu_x, \mu_y, \mu_z)^T||
    \end{bmatrix}^T. \nonumber
\end{align}
Note that the distances between the camera and Gaussian centers are still $||(\mu_x, \mu_y, \mu_z)^T||$. The visual illustration is shown in \Cref{fig:splatting} (a) and (b). This ensures the relative accuracy of the sonar physical model.}

The next step after the approximate projection operation is orthographic projection. The 3D Gaussians are converted to 2D Gaussians with covariance matrix $\varSigma_{\text{2D}}'$ obtained by dropping the last row and column, mathematically, 
\begin{align}
\varSigma_{\text{2D}}' = 
\begin{bmatrix}
1 & 0 & 0\\
0 & 1 & 0
\end{bmatrix}
\varSigma'
\begin{bmatrix}
1 & 0 \\
0 & 1 \\
0 & 0 
\end{bmatrix}
\end{align}
The 2D Gaussians are $\alpha$-blended using \Cref{eq:alpha_blending} to synthesize the camera image.

\subsection{The Missing Cone}
\label{sec:missing_cone}
In small baseline imaging scenarios, information about the covariances ($\sigma_{xz}, \sigma_{yz}$) and variance ($\sigma_{zz}$) associated with the depth-axis and the mean ($\mu_z$) along the depth-axis are not captured by the camera images. 

To illustrate this, consider a simple volumetric imaging scenario in \Cref{fig:fusion_rationale}. 
We consider 2D volume (the arguments are extendable to 3D volume) and assume no inter-reflections and far-field sensing modalities, i.e., volumetric scene after the local affine approximation in the GS algorithm.
For these volumetric scenes, the rendered RGB image will be the $x$-axis projection (splat) of the volume. 

Using the Fourier-slice theorem, the Fourier transform of the camera image is the same as the x-slice of the volume's Fourier transform.  By taking several images from various angles, we are capturing various slices of the volume's Fourier transform. Unfortunately, for the small baseline scenarios~\cite{qadri2024aoneus}, we have the missing cone problem~\cite{liu2019analysis, macias1988missing, delaney1998globally}, which refers to a cone in the Fourier space that was not sensed by the camera images. This limits the fidelity of the 3D reconstructions due to missing frequency information. 

Our approach to solving this problem is to utilize complementary information present in time-resolved measurements, particularly time-resolved measurements from sonar, which can be obtained by taking the z-axis projection of the volume as we will discuss next in (\Cref{sec:proposed}). From the Fourier-slice theorem, this is equivalent to obtaining a z-slice of the volume's Fourier transform as shown in \Cref{fig:fusion_rationale}(c). Hence, sonars can help capture the missing cone and augment camera data, particularly when camera baselines are limited.

%% file: sections/04_proposed_method.tex
\section{Z-Axis Gaussian Splatting for Camera-Sonar Fusion}
\label{sec:proposed}
We will first extend the Gaussian splatting model to two commonly used sonars, i.e. echosounder and FLS, by exploiting the physics of how the sonars operate. After that, we will combine the camera and sonar Gaussian splatting to build a fusion algorithm capable of reconstructing the 3D Gaussian parameters accurately even for small baseline imaging scenarios. 





\subsection{Echosounder}
\label{sec:echosonar_splatting}
A single-beam echosounder utilizes a single transducer element to emit a sound pulse toward the target area and captures the echo, a time-varying sound pulse that bounces back from the scene. 
This echoed signal is a function of acoustic reflectivity and depth of various scene points. 
For simplicity, we assume the acoustic albedo is constant for all the objects in the scene.
Therefore, the echoed signal can also be interpreted as a 1D histogram of scene depths computed through acoustic time-of-flight. 

In \Cref{fig:splatting}, we illustrate the proposed splatting operation for echosounder. We splat the 3D Gaussians along the $z$-axis, while computing the transmission and alpha values similar to the camera splatting. The covariance of the 1D projection of 3D Gaussians is:
\begin{equation}
    \varSigma_{\text{1D}}' = 
    \begin{bmatrix}
    0 & 0 & 1\\
    \end{bmatrix}
    \varSigma'
    \begin{bmatrix}
    0  \\
    0  \\
    1  
    \end{bmatrix}
    = \sigma_{zz}. 
\end{equation}

Therefore, the echosounder captures the information about mean $\mu_z$ and variance $\sigma_{zz}$ along the depth axis that was missing in the camera data. 
The proposed splatting operation computes the depth histogram of the volumetric scene while accounting for the visibility term accurately. 

\qq{
We show the steps of the $z$-axis splatting for echosounder in \Cref{alg:rendering_echosounder}. 
We splat Gaussians for each $xy$-pixel during rasterization.
Therefore, if the scene is represented as one large Gaussian or several small Gaussians, the rasterized Z-splat result would be same. 
}

\begin{algorithm}[t]
\caption{Echosounder rendering}
\label{alg:rendering_echosounder}
\begin{algorithmic}
    \State $Z[d] = 0;$      
    \For{\texttt{each point on the $xy$-plane}}{
    \For{\texttt{each visible Gaussian Kernel}}
        \For{\texttt{each bin $i$ within $\mu_z \pm 3 \sigma_{zz}$}}
        \State $z = $ \texttt{center of the bin}
        \State $Z[i] += T \cdot \alpha \cdot \exp(-\frac{(z-\mu_z)^2}{2\sigma_{zz}})$
        \EndFor
    \EndFor   
    \EndFor
 }
\end{algorithmic}
\end{algorithm}


\subsection{Forward Looking Sonar (FLS):}
\label{sec:fls_splatting}
\begin{figure}[t]
    \centering
    \setlength{\abovecaptionskip}{-0.12cm}
     \includegraphics[width=\linewidth]{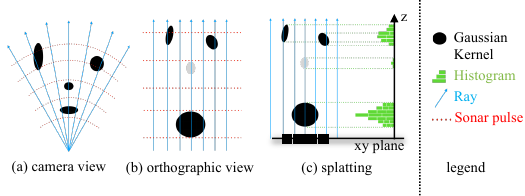}
     \caption{\textbf{Ray View Transformation and Z-Axis Splatting} 
    (a) This illustration shows the camera view. The covariance of Gaussians in the camera view is $\varSigma = W^T \varSigma W$, which transforms the Gaussians from the world view to the camera view.
    (b) The Gaussians are transformed into the ray view through an \qq{local} affine approximation of the projection transform using the Jacobian ($J$). The covariance matrix of the Gaussians will be $\varSigma' = J^T \varSigma J$. 
    (c) The transformed 3D Gaussian is then projected (splat) onto the $xy$-plane for rendering camera and $z$-axis for rendering echosounder (for collocated camera and echosounder). The gray Gaussian is occluded by the Gaussian in the front, so the Transmission($T$) of that Gaussian is smaller than the others independent of whether we are rendering camera or sonar. \qq{Based on \Cref{alg:rendering_echosounder} and \Cref{alg:FLS_rendering}, each ray undergoes splatting independently, ensuring that if a Gaussian is rasterized by multiple rays, it will be splatted multiple times.}
     }
     \label{fig:splatting}
\end{figure}

FLS is similar to an echosounder, except it contains multiple linear transducer elements. 
Through beamforming, FLS recovers the range and azimuth information but not the elevation. Therefore, in the orthographic view (\Cref{fig:splatting}(b)), FLS measures depth histograms for various $y$ values. 

To build a rendering algorithm for FLS, we splat the 3D Gaussians on the $y-z$ plane, resulting in 2D Gaussians with covariance matrix:
\begin{equation}
    \varSigma'_{2D} = 
    \begin{bmatrix}
       \sigma_{yy} & \sigma_{yz} \\
       \sigma_{yz} & \sigma_{zz}
    \end{bmatrix}. 
\end{equation}
We compute the transmission and alpha values similar to camera splatting, as the visibility terms do not change for FLS. We identify the steps in FLS rendering in \Cref{alg:FLS_rendering}.


In practice, an RGB camera often has a higher spatial resolution than a sonar, so the information on the y-axis mainly comes from the RGB camera. 
Therefore, we expect the reconstruction accuracy and quality of novel synthetic views supervised by FLS to be slightly better than those from echosounder, but not significantly better.

\begin{algorithm}[t]
\caption{FLS rendering}\label{alg:FLS_rendering}
\begin{algorithmic}
    \State $Z[h, d] = 0;$      
    \For{\texttt{each point on the $xy$-plane}}{
    \For{\texttt{each visible Gaussian Kernel}}
        \For{\texttt{each bin $i$ within $\mu \pm 3 \cdot \sigma_{zz}$}}
        \State $z = $ \texttt{center of the bin}
        \State $j = $ \texttt{y-center of the Gaussian}
        \State $Z[j, i] += T \cdot \alpha \cdot \exp(-\frac{(z-\mu_z)^2}{2\sigma_{zz}})$
        \EndFor
    \EndFor   
    \EndFor
 }
\end{algorithmic}
\end{algorithm}

\subsection{Fusion of cameras and sonars}
We fuse the camera and sonar information by jointly minimizing the error between the rendered and measured data for both sensors. 
We define the camera loss $\mathcal{L}_c$ as:
\begin{equation}
    \mathcal{L}_c = \|I(x,y) - I_{\gs}(x,y)\|_1,
\end{equation}
where $I(x,y)$ represents the measured camera image and $I_{gs}(x,y)$ is the rendered image from Gaussian splatting. 

The sonar loss ($\mathcal{L}_s$) is defined as
\begin{equation}
    \mathcal{L}_s = 
     \begin{cases}
       \|S(z)-S_{\gs}(z)\|_2, & \text{for Echosounder}\\
       \|S(y, z) - S_{\gs}(y,z)\|_2, & \text{for FLS}. 
    \end{cases}
\end{equation}
Here $S(z)$ and $S(y,z)$ are the measured echosounder and FLS data; $S_{\gs}$ and $S_{\gs}(y,z)$ are the Gaussian splatted renderings discussed in~\Cref{sec:echosonar_splatting} and~\Cref{sec:fls_splatting}.
\qq{Note that $L_s$ is a vector norm for echosounder and a Frobenius norm for FLS.}

We define the loss function as a weighted combination of camera and sonar loss functions: 
\begin{equation}
    \mathcal{L} = \mathcal{L}_c + w \cdot \mathcal{L}_s. 
\end{equation}

Empirically, we found that setting $0.1 \leq w \leq 3 $ is suitable for most scenarios. We can tune $w$ based on the confidence between camera and sonar measurements. 
For instance, in \Cref{fig:simulation_photometric}, we choose $w = 3$ for the living room scene because the scene does not have a lot of texture. We can also choose adaptive weighing schemes similar to Qadri et al.~\cite{qadri2024aoneus} where more importance is given to sonar measurements in the beginning and camera measurements in the later iterations, though we did not find that weighing scheme useful in this case.

\subsection{Implementation}

We have implemented the proposed algorithms by extending the publicly available Gaussian splatting source code shared by Kerbl et al.~\cite{kerbl20233d}. 
We computed the derivatives analytically and wrote cuda kernels to improve the rendering and training speed. The average training time for all the simulated experiments in the paper is around $5$ minutes on a NVIDIA 4090 GPU, while real experiments took $20$ minutes for echosounder and 8 minutes for FLS.

%% file: sections/05_experimental_results.tex
\begin{figure*}[t]
    \centering
    \setlength{\abovecaptionskip}{-0.12cm}
     \includegraphics[width=\textwidth]{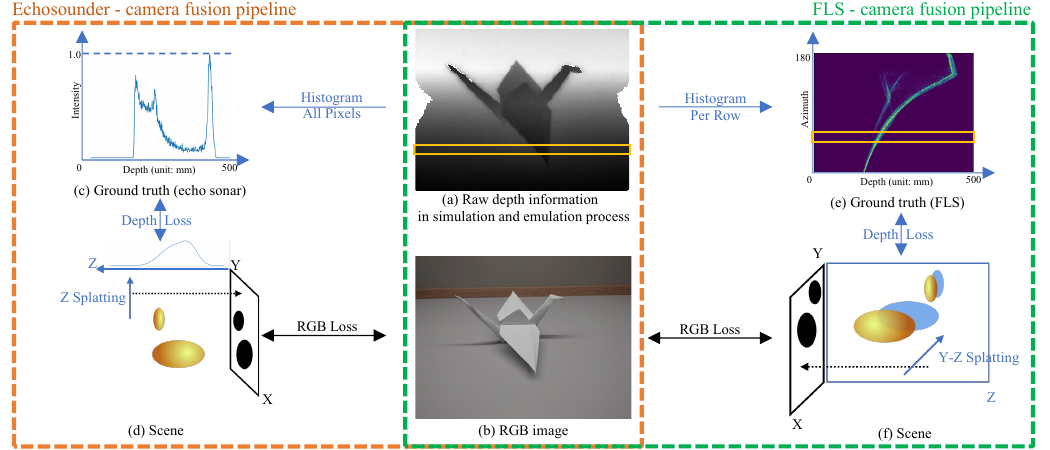}
     \caption{
     \textbf{Simulation and emulation training for both echosounder and FLS fusion techniques} 
    (a) Raw depth image captured with Time-of-Flight (ToF) camera.
    (b) An RGB image captured with a camera.
    (c) Simulated echosounder intensity was generated using the depth histogram and utilized as ground truth during training.
    (d) A 3D Gaussian scene. We use $xy$-splatting to render RGB images and $z$-splatting to render echosounder depth intensity distribution. 
    (e) Simulated FLS intensity generated by histogramming depth per row. 
    (f) A 3D Gaussian scene, and we splat along $xy$-direction to render RGB image and along $yz$-direction to render FLS image. 
    We minimize the sum of RGB loss and corresponding depth loss to train the camera-sonar fusion algorithms.
    }
    \vspace{-0.1in}
     \label{fig:echo_sonar_simulation_training_process}
\end{figure*}

\section{Experimental Results}

In this section, we systematically evaluate the proposed fusion techniques using simulated, emulated, and hardware-capture data. Novel view synthesis refers to generating images from camera views not in the training data. Using standard photometric metrics (PSNR, SSIM, and LPIPS), we compare the novel views generated by traditional Gaussian splatting and the proposed fusion techniques. We also compare the reconstructed geometry as 3D point clouds using Chamfer distance, F1 score, precision, and recall.




\vspace{-0.05in}

\subsection{Simulation results}
\label{sec:simulation}
We use Mitsuba~\cite{nimier2019mitsuba} to generate the simulated datasets. For each camera and sonar view, we render the image from the camera view and the depth map from the sonar view. We simulate echosounder intensity by histogramming the depth values of the entire scene, resulting in an intensity distribution across depth that mimics the transient sonar measurements. This histogram serves as the ground truth either for training or for testing. 
The left part of \Cref{fig:echo_sonar_simulation_training_process} illustrates the process of generating the ground truth echosounder intensity and the predicted echosounder intensity using the proposed method. 

To simulate FLS data, we histogram the depth values per row rather than the whole image. Each row represents one azimuth angle $\theta$ in the FLS. 
The resolution of the histogram along the z-axis is determined by the range $r$ that the FLS can resolve. The right part of \Cref{fig:echo_sonar_simulation_training_process} illustrates the process of generating ground-truth FLS scans.
To initialize Gaussians, we found that random initialization to be more reliable than COLMAP initialization, as the latter sometimes suffered due to small baselines in our datasets. 

\begin{figure*}[t]
    \centering
    \setlength{\abovecaptionskip}{-0.12cm}
     \includegraphics[width=\textwidth]{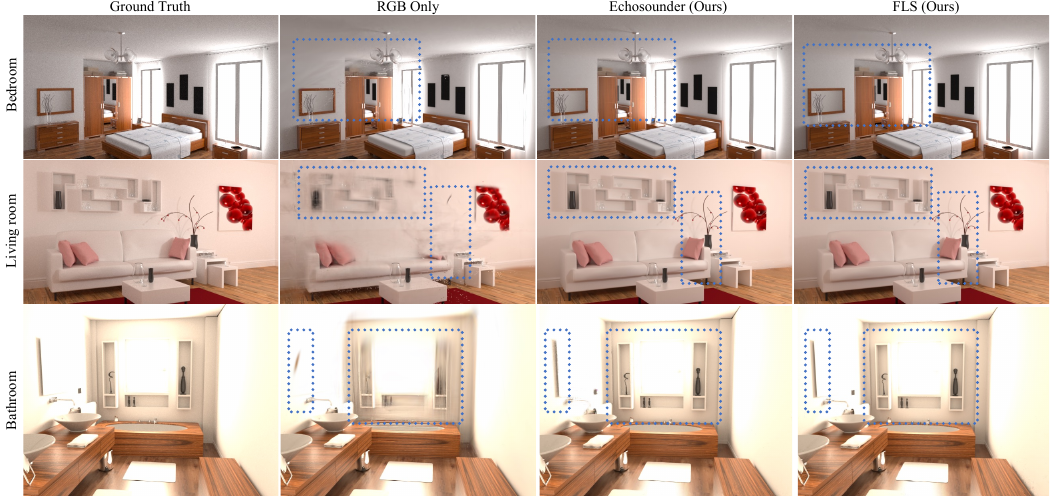}
     \caption{
     \textbf{Novel view synthesis comparison:} The incorporation of depth information notably mitigates the presence of floaters in the reconstructed scene. Moreover, depth information accurately positions the Gaussian kernels, particularly in scenes with uniform color or overexposure. 
     The average SSIM, PSNR, and LPIPS metrics for the entire test set comprising 263 novel views are presented in \Cref{tab:simulated_data_photometric_complex}.
    }
     \label{fig:simulation_photometric}
     \vspace{-0.2in}
\end{figure*}

We provide two different types of scenes in the simulation process: (i) Room-sized scenes from Mitsuba Gallery (\Cref{fig:simulation_photometric}) and (ii) Single-object scenes from Mitsuba Gallery, Stanford 3D Scanning Repository and online free resources.
In the room-sized scenes shown in \Cref{fig:simulation_photometric}, we cannot freely maneuver the camera around the scene or object and obtain a full 360-degree scan due to the geometric constraints, and hence, we are in a restricted baseline scenario. Further, the scenes contain saturated regions due to windows and little texture. For these imaging scenarios, the camera-based reconstruction is ill-posed, and the addition of depth information better constrains the reconstruction problem. 


We present the novel view synthesis comparisons (photometric comparisons) in \Cref{tab:simulated_data_geometry_complex} and geometry reconstruction comparisons in \Cref{tab:simulated_data_photometric_complex}. 
For novel view synthesis, we outperform the RGB-only method on PSNR, SSIM, and LPIPS metrics. Specifically, we observe an average 5~dB improvement in PSNR compared to camera-only GS algorithms and a 10~dB increase on challenging scenes that contain little texture (living room scene). 

The reconstructed 3D Gaussians typically have a small variance (of a couple of pixels) and, hence, can act as reconstructed point clouds. 
We tessellated the ground truth mesh and considered the mesh vertices as the ground truth point cloud. 
This tessellation allows us to compute various metrics (Chamfer distance, Recall, precision, F1-score) between the reconstructed point cloud and the ground truth. 
In \Cref{tab:simulated_data_photometric_complex}, we show the comparisons across all the methods. The proposed techniques consistently outperform the traditional GS algorithms. 
we observe an average 50\% improvement in Chamfer distance and a tenfold increase in the scenes with little texture (living room scene).


The primary reason for these improvements lies in our method's effective mitigation of the issue of floaters in the scene, a common challenge encountered in volumetric rendering methods. By reducing erroneous placement of Gaussians and enhancing both geometry accuracy and photometric details, we achieve significant advancements in overall performance. The results also indicate that FLS outperforms echosounder, albeit with only a slight improvement. This confirms our hypothesis from \Cref{sec:fls_splatting} that FLS will provide little additional information in the $y$-axis, as cameras also provide information along that dimension and at a higher resolution. 

\begin{table*}[!t]
    \renewcommand{\arraystretch}{1.3}
    \setlength{\abovecaptionskip}{0cm}
    \setlength{\belowcaptionskip}{-0.2cm}
    \caption{Simulation: Novel view synthesis comparisons (Room sized scenes)}
    \label{tab:simulated_data_photometric_complex}
    \centering
    \begin{tabular}{c||c|c|c||c|c|c||c|c|c}
    \hline
    & \multicolumn{3}{c||}{RGB Only} & \multicolumn{3}{c||}{Echosounder (Ours)} & \multicolumn{3}{c}{FLS (Ours)}\\
    \hline
    Scene & PSNR$^{\uparrow}$ & SSIM$^{\uparrow}$ & LPIPS$^{\downarrow}$ 
        & PSNR$^{\uparrow}$ & SSIM$^{\uparrow}$ & LPIPS$^{\downarrow}$ 
        & PSNR$^{\uparrow}$ & SSIM$^{\uparrow}$ & LPIPS$^{\downarrow}$\\
    \hline\hline
    Bedroom     & \textcolor{red}{31.855}             & \textcolor{red}{0.881}          & \textcolor{red}{0.242} 
                & 35.264                              &\textcolor{mydarkgreen}{0.893}         & 0.205 
                & \textcolor{mydarkgreen}{35.348}           & 0.891                           & \textcolor{mydarkgreen}{0.177}\\
    \hline
    Living room & \textcolor{red}{27.508}           & \textcolor{red}{0.874}            & \textcolor{red}{0.331} 
                & 37.790                            & 0.948                             & 0.182 
                & \textcolor{mydarkgreen}{38.457}         & \textcolor{mydarkgreen}{0.951}          & \textcolor{mydarkgreen}{0.176}\\
    \hline
    Bathroom    & \textcolor{red}{27.465}          & \textcolor{red}{0.872}             & \textcolor{red}{0.177}
                & 33.381                           & 0.929                              & 0.092
                & \textcolor{mydarkgreen}{35.753}        & \textcolor{mydarkgreen}{0.952}           & \textcolor{mydarkgreen}{0.073}\\
    \hline
    \end{tabular}
    \begin{tablenotes}  
        \item \textbf{PSNR, SSIM, and LPIPS metrics.} We use PSNR, SSIM, and LPIPS to evaluate the quality of the predicted novel view RGB images. \textcolor{mydarkgreen}{The best results are in green}, \textcolor{red}{while the worst are in red}. We can see that our methods outperform the RGB-only method in all scenes. While the FLS method has the best performance, as expected, the improvement compared to the echosounder is small. 
    \end{tablenotes}  
\end{table*}

\begin{table*}[!t]
    \renewcommand{\arraystretch}{1.3}
    \setlength{\abovecaptionskip}{0cm}
    \setlength{\belowcaptionskip}{-0.2cm}
    \setlength{\tabcolsep}{4.1pt}
    \caption{Simulation: Geometric comparisons (Room sized scenes)}
    \label{tab:simulated_data_geometry_complex}
    \centering
    \begin{tabular}{c||c|c|c|c||c|c|c|c||c|c|c|c}
    \hline
    & \multicolumn{4}{c||}{RGB Only} & \multicolumn{4}{c||}{Echosounder (Ours)} & \multicolumn{4}{c}{FLS (Ours)}\\
    \hline
    Scene &Chamfer$^{\downarrow}$&Precision$^{\uparrow}$&Recall$^{\uparrow}$&F1$^{\uparrow}$
          &Chamfer$^{\downarrow}$&Precision$^{\uparrow}$&Recall$^{\uparrow}$&F1$^{\uparrow}$
          &Chamfer$^{\downarrow}$& Precision$^{\uparrow}$&Recall$^{\uparrow}$&F1$^{\uparrow}$\\
    \hline\hline
    Bedroom & \textcolor{red}{0.374}   & \textcolor{red}{0.893}     & \textcolor{red}{0.549}    & \textcolor{red}{0.680} 
            & \textcolor{mydarkgreen}{0.163} & 0.997                      & \textcolor{mydarkgreen}{0.667}  & \textcolor{mydarkgreen}{0.799}
            & 0.198                    & \textcolor{mydarkgreen}{0.998}   & 0.622                     & 0.767\\
    \hline
    Living Room & \textcolor{red}{3.382}   & \textcolor{red}{0.825}   & \textcolor{red}{0.084}      & \textcolor{red}{0.152}
                & \textcolor{mydarkgreen}{0.291} & 0.977                    & 0.512                       & 0.672
                & 0.359                    & \textcolor{mydarkgreen}{0.998} & \textcolor{mydarkgreen}{0.540}    & \textcolor{mydarkgreen}{0.701}\\
    \hline
    Bathroom    &  \textcolor{red}{4.545}   &  \textcolor{red}{0.616}   & \textcolor{mydarkgreen}{0.594}   &  \textcolor{red}{0.605}
                &  4.136                    &   0.882                   &      0.521                 &  \textcolor{mydarkgreen}{0.655}
                &  \textcolor{mydarkgreen}{3.912} &  \textcolor{mydarkgreen}{0.907} &       0.485                & 0.632\\
    \hline
    \end{tabular}
    \begin{tablenotes}
        \item \textbf{Chamfer, Precision, Recall, and F1 score metrics.} We use Chamfer, Precision, Recall, and F1 score to evaluate the quality of the predicted 3D Gaussian splatting.\textcolor{mydarkgreen}{The best results are in green}, \textcolor{red}{while the worst are in red}. The metrics are calculated by comparing the point cloud of the predicted Gaussian splatting with the ground truth mesh vertices. Our methods outperform the RGB only method in most metrics, and the FLS method has the best performance. 
    \end{tablenotes}
    \setlength{\tabcolsep}{6pt}
\end{table*}

\begin{figure*}[t]
    \centering
    \setlength{\abovecaptionskip}{-0.12cm}
     \includegraphics[width=\textwidth]{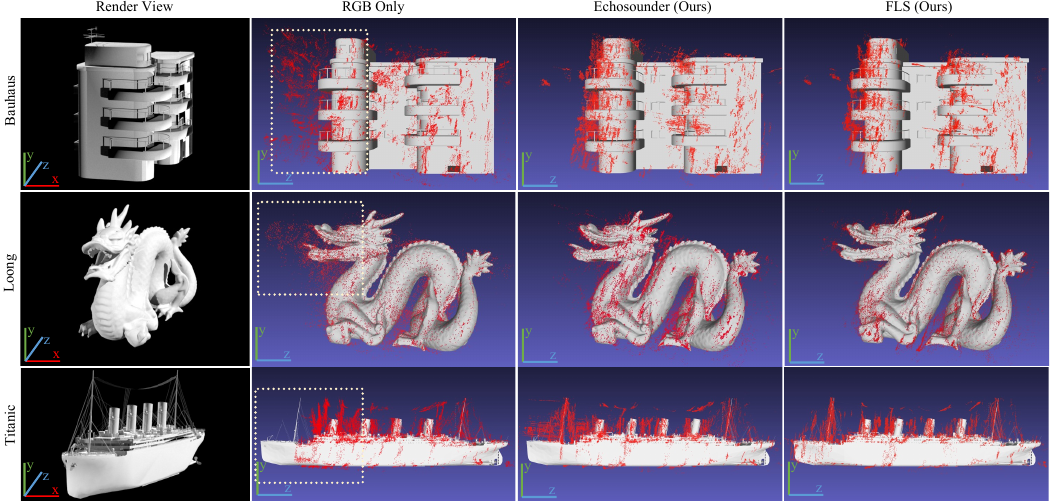}
     \caption{
     \textbf{Geometry comparison on one-object scenes.} We captured the data by moving the camera only along the $x$-axis. 
     We show ground truth meshes and superimpose the reconstructed Gaussians as point clouds. 
     In the highlighted regions, we can observe that camera-only methods reconstruct the geometry inaccurately along the $z$-axis, whereas the proposed fusion techniques reconstruct the geometry accurately. 
    }
     \label{fig:simulation_geometry}
\end{figure*}

In the second set of scenes, instead of complex scenes with multiple objects, we created a series of scenes with only one diffuse object without texture. 
Similar to Qadri et al.~\cite{qadri2024aoneus}, we restricted camera and sonar movement along the $x$-axis within a small range (2.3 degree on the circle) across these simplified scenes. We visualize the results in \Cref{fig:simulation_geometry}, showcasing the ground truth as a mesh and the predicted Gaussian kernels as red points.
In \Cref{tab:simulated_data_geometry_one_object}, we quantify the accuracy of reconstructed geometry with various metrics. We can observe that the proposed fusion consistently performs better than camera-only techniques. 

\qq{\textbf{Comparison to ToRF~\cite{attal2021torf}.} We found that the camera-only GS method outperforms ToRF. GS achieves a PSNR of 33.91 dB when trained solely on RGB images from the ToRF bedroom dataset, and 36.42 dB with fusion, compared to ToRF's PSNR of 29.79 dB with fusion. Note that the PSNR for ToRF is taken from its original paper.}

\begin{table*}[!t]
    \renewcommand{\arraystretch}{1.3}
    \setlength{\abovecaptionskip}{0cm}
    \setlength{\belowcaptionskip}{-0.2cm}
    \setlength{\tabcolsep}{4.8pt}
    \caption{Simulation: Geometric comparisons (One-object scenes)}
    \label{tab:simulated_data_geometry_one_object}
    \centering
    \begin{tabular}{c||c|c|c|c||c|c|c|c||c|c|c|c}
    \hline
    & \multicolumn{4}{c||}{RGB Only} & \multicolumn{4}{c||}{Echosounder (Ours)} & \multicolumn{4}{c}{FLS (Ours)}\\
    \hline
    Scene &Chamfer$^{\downarrow}$&Precision$^{\uparrow}$&Recall$^{\uparrow}$&F1$^{\uparrow}$
          &Chamfer$^{\downarrow}$&Precision$^{\uparrow}$&Recall$^{\uparrow}$&F1$^{\uparrow}$
          &Chamfer$^{\downarrow}$& Precision$^{\uparrow}$&Recall$^{\uparrow}$&F1$^{\uparrow}$\\
    \hline\hline
    Bauhaus & \textcolor{red}{0.184}   & \textcolor{mydarkgreen}{0.943} & \textcolor{red}{0.693}   & \textcolor{red}{0.799} 
            & 0.134                    & 0.899                    & 0.729                    & 0.805 
            & \textcolor{mydarkgreen}{0.127} & \textcolor{red}{0.894}   & \textcolor{mydarkgreen}{0.772} & \textcolor{mydarkgreen}{0.829}\\
    \hline
    Loong   & \textcolor{red}{0.0184}   & \textcolor{red}{0.8462}    & \textcolor{red}{0.916}   & \textcolor{red}{0.846} 
            & 0.0093                    & 0.8627                     & \textcolor{mydarkgreen}{0.944} & 0.902 
            & \textcolor{mydarkgreen}{0.0086} &  \textcolor{mydarkgreen}{0.8815} & 0.938                    & \textcolor{mydarkgreen}{0.909}\\
    \hline
    Titanic & \textcolor{red}{0.164}   & \textcolor{red}{0.949}   & \textcolor{red}{0.808}   & \textcolor{red}{0.873} 
            & 0.103                    & \textcolor{mydarkgreen}{0.997} & 0.887                    & 0.939 
            & \textcolor{mydarkgreen}{0.069} & 0.984                    & \textcolor{mydarkgreen}{0.932} & \textcolor{mydarkgreen}{0.957}\\
    \hline
    \end{tabular}
    \begin{tablenotes}
        \item \textbf{Chamfer, Precision, Recall, and F1 score metrics.} We use Chamfer, Precision, Recall, and F1 score to evaluate the quality of the predicted 3D Gaussian splatting. \textcolor{mydarkgreen}{The best results are in green}, \textcolor{red}{while the worst are in red}. These metrics are computed by comparing the point cloud of the predicted Gaussian splatting with the ground truth mesh vertices. Our methods consistently outperform the RGB-only approach across most metrics, and FLS exhibits slightly better performance than echosounder.
    \end{tablenotes}
    \setlength{\tabcolsep}{6pt}
\end{table*}

\vspace{-0.05in}
\subsection{Hardware emulation results}

\begin{figure*}[t]
    \centering
    \setlength{\abovecaptionskip}{-0.3cm}
     \includegraphics[width=\textwidth]{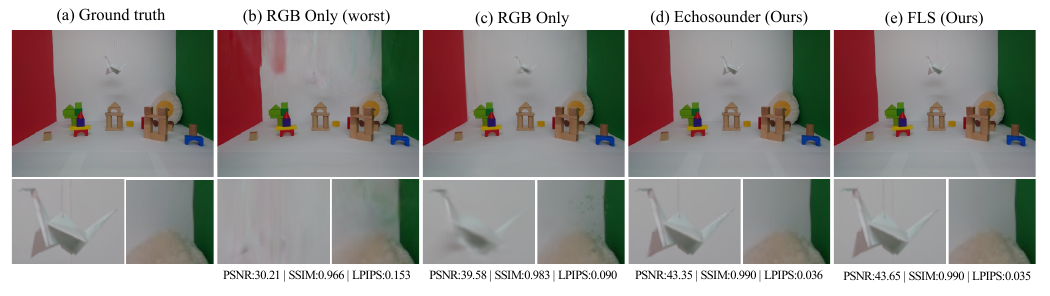}
     \caption{
     \textbf{Novel view synthesis comparison on emulated hardware.} We set up a Cornell box in the lab, captured both RGB and depth images, and emulated the echosounder and FLS data. 
     In the reconstructed scene, all methods work well on the objects with high-contrast textures. However, the RGB-only technique fails to reconstruct the white object with the same color as the background and also suffers from color bleeding.
     Our methods, on the other hand, successfully reconstruct the white object and do not suffer from color bleeding. 
     For different random seeds, RGB-only techniques have high variance in the reconstructed results and have poor reconstructions (b), while our methods have consistent performance. 
     We plot the variance across different runs in \Cref{fig:emulation_variance}, and we can observe that our techniques are robust to random initial seeds. 
    }
     \label{fig:emulation_cornell_box}
\end{figure*}

\begin{table}[!t]
    \renewcommand{\arraystretch}{1.3}
    \setlength{\abovecaptionskip}{0cm}
    \setlength{\belowcaptionskip}{-0.2cm}
    \caption{Emulation Performance: Cornell Box}
    \label{tab:cornell_box_photo}
    \centering
    \begin{tabular}{c||c|c|c}
    \hline
     & PSNR$^{\uparrow}$ & SSIM$^{\uparrow}$ & LPIPS$^{\downarrow}$\\
    \hline\hline
    RGB Only & 37.499 & 0.977 & 0.093 \\
    \hline
    Echosounder (Ours) & 42.089 & 0.987 & 0.037\\
    \hline
    FLS (Ours) & \textbf{42.142} & \textbf{0.988} & \textbf{0.036}\\
    \hline
    \end{tabular}
\end{table}

In addition to the simulation, we emulate the sonars using the camera and Lidar on the Sony Xperia II smartphone. 
We have built a Cornell box scene with multiple objects in our lab. We translate the phone inch-by-inch on the $xy$-plane to obtain training data. 
We capture RGB images and pixel-wise depth measurements simultaneously, mirroring the process employed during simulation in \Cref{sec:simulation}. 
We position the phone in between the training samples to generate the test dataset. We extract camera poses and Structure-from-Motion (SFM) points using COLMAP and train the three models of GS, Echosounder, and FLS. 

The test results are shown in \Cref{fig:emulation_cornell_box} and quantified in \Cref{tab:cornell_box_photo}.
From \Cref{fig:emulation_cornell_box}, it is evident that the RGB-only reconstruction results in significantly more blur on the white crane, box edges, and egg doll. 
This could be attributed to COLMAP creating denser points on clear edges but not on the background or white objects that share the same color as the background. 
During optimization, the RGB-only method would not produce more Gaussians because of the similar color. 
However, our methods surpass the RGB-only approach by producing clear and sharp edges with the aid of $z$-axis information. The average PSNR of our methods is 5~dB higher than that of RGB-only, and our techniques also have higher SSIM and lower LPIPS, as shown in \Cref{tab:cornell_box_photo}. 


\begin{figure}[t]
    \centering
    \setlength{\abovecaptionskip}{-0.3cm}
     \includegraphics[width=0.75\linewidth]{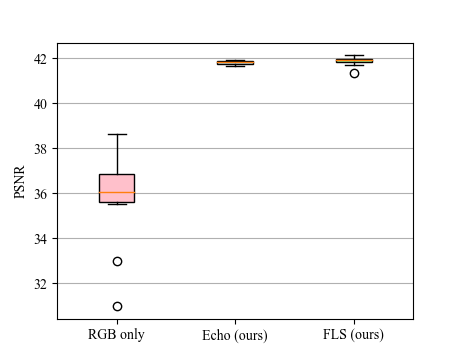}
     \caption{\textbf{Variance comparison for emulations.} We ran the GS algorithms 10 times with the same training and test datasets but with different random seeds. 
     We can see that the variance of our methods is much lower than the RGB-only method. 
     This indicates that our methods have consistent performance, while the RGB-only method can sometimes result in poor results. Note that even the best results with RGB-only are worse than the proposed techniques.
     }
     \label{fig:emulation_variance}
\end{figure}

During the experiments, we observed that the RGB-only method exhibits high variability. To investigate further, we conducted 10 training runs for each model using the same training dataset, test dataset, and initialized SFM points, but with random initial seeds. 
In ~\Cref{fig:emulation_variance}, we show the box and whisker plot to visualize the variance between the runs. 
The RGB-only method can perform poorly at times and exhibits significant variance across each training process. In contrast, both of our methods are robust and have consistent performance. Furthermore, even our worst-performing run outperforms the RGB-only method.

\vspace{-0.05in}
\subsection{Hardware results - Echosounder}

For our set of real hardware experiments, we first utilize our proposed method for sensor fusion between RGB cameras and a monostatic active sonar to capture acoustic ToF data. Echo sounding is the technique of using an active sonar to determine depth/range.
Thus we have built a prototype of an echosounder using a single speaker and microphone that is attached to a DSLR camera for collecting real acoustic ToF data along with RGB images. We describe our setup and data processing procedures below. 


\begin{figure}[t]
    \centering
    \setlength{\abovecaptionskip}{-0.12cm}
    \includegraphics[width=0.8\columnwidth]{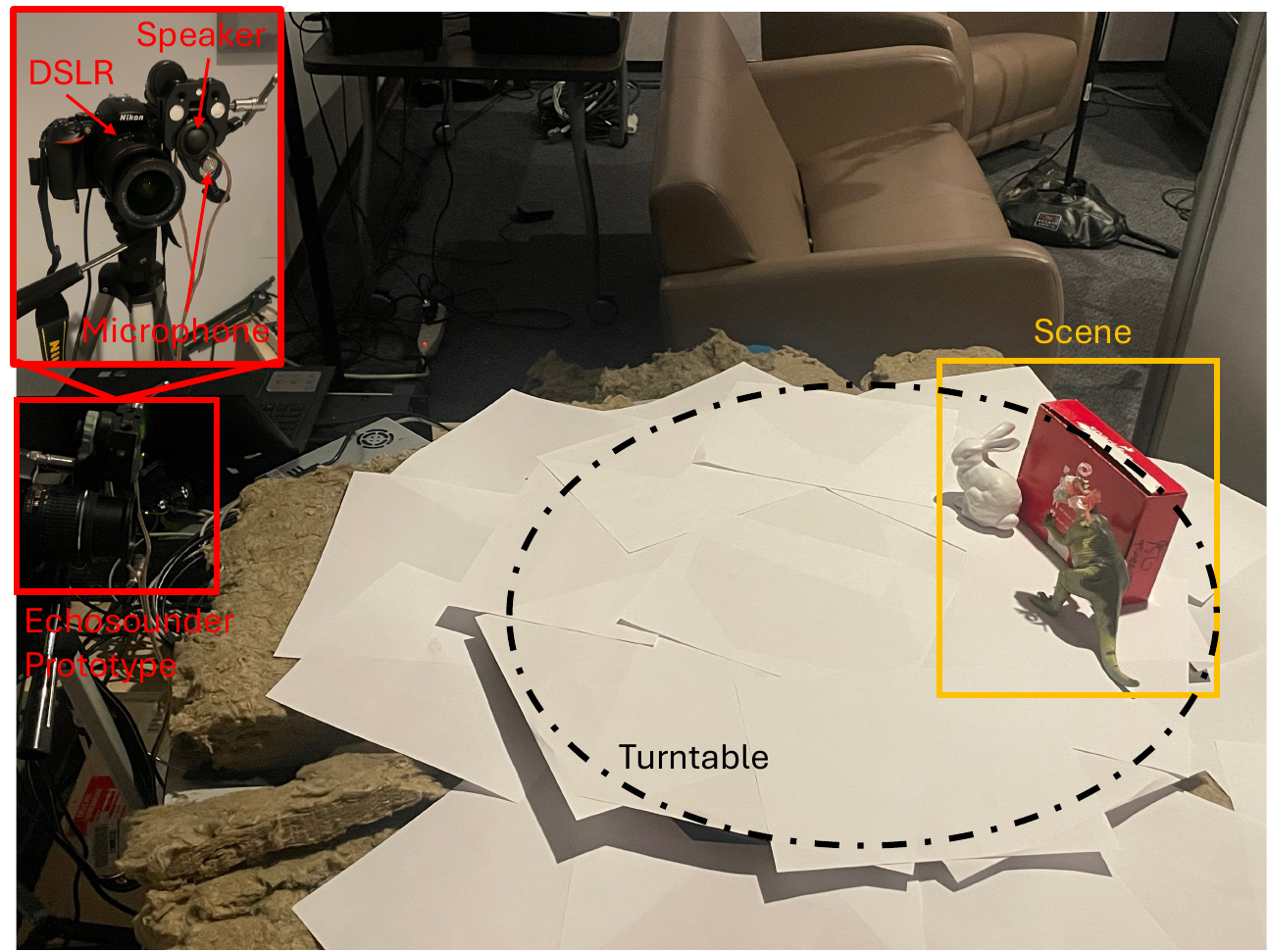}
     \caption{\textbf{ 
     Hardware prototype} consists of a DSLR camera with speaker and microphone. It is used to image a scene on the motorized turntable.
    }
     \label{fig:airsassetup}
\end{figure}

\begin{figure}[t]
    \centering
    \setlength{\abovecaptionskip}{-0.12cm}
     \includegraphics[width=\linewidth]{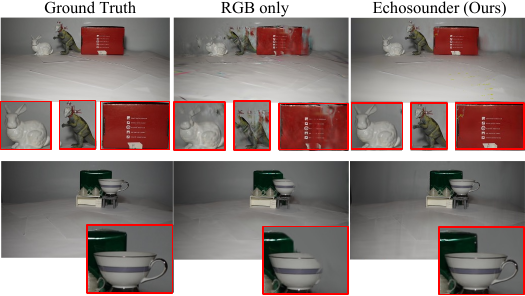}
     \caption{
    \textbf{Qualitative comparison of echosounder real-data results.} The comparison presents two scenes captured using a DSLR camera and echo-sonar with a turntable setup. Our method demonstrates a noticeable improvement in performance over the baseline RGB-only method, both quantitatively and qualitatively.
    }
     \label{fig:echo_sonar_real_data}
\end{figure}


\textbf{Experimental setup and data capture:} Our setup, visualized in Figure~\ref{fig:airsassetup} and similar to other circular in-air systems~\cite{Blanford2019,park2020alternative,reed2022sinr}, consists of (1) acoustic transducer array consisting of loudspeaker tweeter (Peerless by Tymphany OX20SC02-04) and a microphone (GRAS 46AM); (2) a Nikon D5600 DSLR camera; and (3) a motorized circular platform (Rotary Table RTLA-90-200M) to rotate targets. As seen in the figure, white paper and a white board was used to cover the turntable and background, otherwise parts of the scene would be static while the objects moved, not reflecting the experimental scenario of a moving sonar. The tweeter transmits a linear frequency modulated chirp with peak voltage of 3V for a duration of 1 ms waveform from starting frequency $10 kHz$ to stopping frequency $30 kHz$ with a sampling frequency of $100 kHz$. 


For data capture, we rotate the target scene at 1-degree increments from 0 to 40 degrees while simultaneously capturing RGB video frames and the acoustic signal from the microphone. We first use COLMAP to perform SfM to obtain an initial point cloud and camera pose. We perform model alignment for SFM points using the information from the experimental setup (radius of the turn table and incremental angle) to roughly align the SFM points in 3D space with the transient histogram to be computed. For the audio signal, we perform similar processing for matched filtering by Reed et al.~\cite{reed2022sinr,reed2023neural}: (1) we implement group delay correction in the frequency domain; (2) we subtract background measurement of the scene with no object; (3) we convert the signal into the analytic domain using the Hilbert Transform; (3) we perform matched filtering in the frequency domain, and then (4) take the magnitude of the complex signal to form a transient histogram. 

For initialization of the Gaussians, we utilize the inital SfM point cloud generated from COLMAP as well as include additional 3D Gaussians ($N=10,000$ for our experiments) who are placed at $(x,y,z)$ locations corresponding to maximum energy in the transient histogram. This greatly improved the performance of the method in practice compared to the baseline Gaussian splatting. 

\textbf{Main experimental results:}  In Fig.~\ref{fig:echo_sonar_real_data}, we show a qualitative comparison of our method compared to an RGB only 3D reconstruction. This scene is a difficult scene as COLMAP SfM points are noisy and inaccurate, which contributes to errors in the RGB only Gaussian-splatting reconstruction. In contrast, our method better resolves the objects in the scene as well as the background, leveraging the transient information from the echosounder. In Tab.~\ref{tab:Sonar_scene_comparison}, we see an improvement in PSNR/SSIM/LPIPS for test views. 

\begin{table}[!t]
    \renewcommand{\arraystretch}{1.3}
    \setlength{\abovecaptionskip}{0cm}
    \setlength{\belowcaptionskip}{-0.2cm}
    \setlength{\tabcolsep}{4.1pt}
    \caption{Echo-Sonar Performance}
    \label{tab:Sonar_scene_comparison}
    \centering
    \begin{tabular}{c||c|c|c||c|c|c}
    \hline
    & \multicolumn{3}{c||}{RGB Only} & \multicolumn{3}{c}{Echosounder (Ours)}\\
    \hline
    Scene & PSNR$^{\uparrow}$ & SSIM$^{\uparrow}$ & LPIPS$^{\downarrow}$ 
        & PSNR$^{\uparrow}$ & SSIM$^{\uparrow}$ & LPIPS$^{\downarrow}$ \\
    \hline\hline
    Rabbit     & 25.882           & 0.874         & 0.282
                & \textbf{30.785}                         &\textbf{0.940}         & \textbf{0.237} \\
    \hline
    Teapot & 39.664         & 0.982            & 0.138 
                & \textbf{39.717}                            & \textbf{0.985}                            & \textbf{0.114} \\
    \hline
    \end{tabular}
    \setlength{\tabcolsep}{6pt}
\end{table}


\qq{\textbf{Effect of baseline:}
We designed the technique for small baselines (small shifts in the camera positions) where sonar provides the most information in the missing cone (\Cref{sec:missing_cone}) and showed that the proposed technique results in higher 3D reconstruction accuracy compared to RGB case. 
As the baseline increases, the missing cone becomes narrower; hence, we will only have diminishing gains from the sonar. 
Here, we quantify the effect of the baseline on the reconstruction performance. 
For this, we rotate the circular table by various ranges of angles and compare the geometric reconstructions between RGB only and the proposed fusion techniques. 
We show the visual reconstruction comparisons in \Cref{fig:baseline} and quantify them in \Cref{tab:baseline_variation_real_data}. We observe that fusion techniques have a significant advantage at small baselines (20$^\circ$), achieving a 24.5\% lower Chamfer distance, but a relatively lower advantage at large baselines (180$^\circ$), with only a 7.9\% decrease in Chamfer distance.
}

\begin{table}[ht]
    \renewcommand{\arraystretch}{1.3}
    \setlength{\abovecaptionskip}{0cm}
    \setlength{\belowcaptionskip}{-0.2cm}
    \setlength{\tabcolsep}{4.1pt}
    \caption{Various Baselines Results}
    \label{tab:real_data_geometry_complex}
    \centering
    \begin{tabular}{c|c|c|c}
        \hline
        Baseline & Metrics & RGB only & Ours echosounder\\
        \hline
         $20^{\circ}$ & Chamfer $\downarrow$ & 0.01455 & 0.01098 \\
          & F1 $\uparrow$ & 0.9534 & 0.9910 \\
          \hline
         $40^{\circ}$ & Chamfer $\downarrow$ & 0.00845 & 0.00656 \\
          & F1 $\uparrow$ & 0.9903 & 0.9962 \\
          \hline
         $90^{\circ}$ & Chamfer $\downarrow$ & 0.0042 & 0.0039 \\
          & F1 $\uparrow$ & 0.9985 & 0.9989 \\
          \hline
         $180^{\circ}$ & Chamfer $\downarrow$ & 0.00366 & 0.00337 \\
          & F1 $\uparrow$ & 0.9994 & 0.9997 \\
          \hline
    \end{tabular}
    \label{tab:baseline_variation_real_data}
\end{table}

\begin{figure}[t]
    \centering
    \setlength{\abovecaptionskip}{-0.12cm}
     \includegraphics[width=\linewidth]{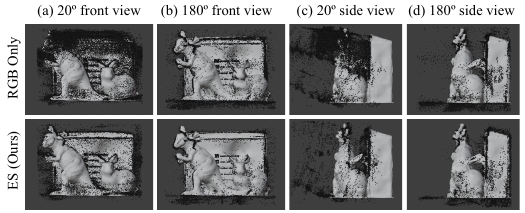}
     \caption{ 
     \qq{\textbf{Effect of baseline.} We visualize two extreme baselines: $20^{\circ}$ and $180^{\circ}$. The ground truth meshes captured with LIDAR are gray, and the reconstructed Gaussians are represented as black point clouds. In both baselines, our method demonstrates better alignment to the underlying geometry and less error in the \( z \) axis, resulting in fewer Gaussians and clearer representations. As the baseline increases, our method's advantage over the RGB-only approach diminishes but still remains positive.}
     }
     \label{fig:baseline}
\end{figure}

\subsection{Hardware results - FLS }
\begin{figure}[!ht]
    \centering
    \setlength{\abovecaptionskip}{-0.12cm}
     \includegraphics[width=0.65\linewidth]{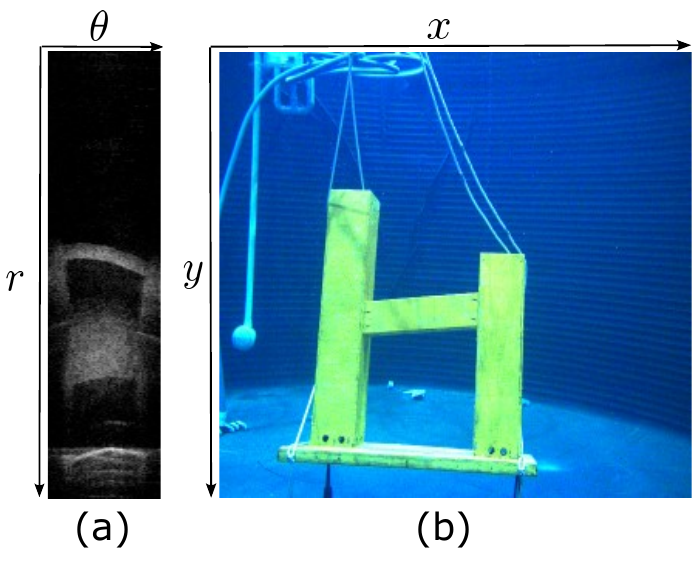}
     \caption{\textbf{Underwater FLS data.} (a) Example sonar measurement: the range $r$ and azimuth $\theta$ are resolved but the elevation angle $\phi$ remains ambiguous. (b) Sample camera image of the submerged test structure.
     }
     \label{fig:flsSonCAM}
\end{figure}

We additionally use our algorithm to fuse measurements from an optical camera and an imaging sonar (FLS) to reconstruct a real object submerged underwater in a water test tank. Acoustic FLS images (example in fig.\ref{fig:flsSonCAM}a ) resolve both the range and azimuth of the reflecting object but not the elevation angle which remains ambiguous. In other words, one can view each column of an imaging sonar image as capturing the transient at a particular azimuth angle which makes this data easily integrable with our method.  

We use an existing dataset containing both imaging sonar measurements (with $14^\circ$ elevation) captured using a Didson imaging sonar mounted on a Bluefin Autonomous Underwater Vehicle and optical camera images captured using a FLIR camera. See papers \cite{qadri2023neural, qadri2024aoneus} for more detail regarding the data and hardware setup. 
\qq{Note that for training, we use all available measurements in the dataset while AONeuS \cite{qadri2024aoneus} only uses specific subsamples.}


\qq{Due to the scattering in water, GS is not suitable for underwater photometric reconstruction, which could be addressed in future work. However, we compare geometric reconstructions.} For this, we first filter out all points outside a bounding box around the object. Then, we align the resulting point cloud with the ground truth model of the target object using the Iterative Closest Point (ICP) algorithm.

We observe from \Cref{flstable1} that integrating FLS information with camera images results in higher reconstruction accuracy, as showcased by the Chamfer distance/precision/recall/F1 metrics (threshold 0.05). This can also be observed quantitatively in \Cref{fig:flsreal}, which shows that when only using optical cameras, the resulting reconstruction contains high errors along the depth axis. On the other hand, integrating FLS information allows our algorithm to better resolve depth using the additional range information. We note that, although our result is slightly worse compared to AONeuS \cite{qadri2024aoneus} (although do expect better performance with more thorough parameter tuning), the runtime of our algorithm is $\sim 9.95 \times$ better.

\begin{table}[!t]
    \renewcommand{\arraystretch}{1.3}
    \setlength{\abovecaptionskip}{0cm}
    \setlength{\belowcaptionskip}{-0.2cm}
    \caption{Hardware FLS: Geometric Metrics}
    \label{flstable1}
    \centering
    \begin{tabular}{c||c|c|c|c}
    \hline
    & Chamfer $^{\downarrow}$ & Precision$^{\uparrow}$ & Recall$^{\uparrow} $  & F1$^{\uparrow}$ \\
    \hline\hline
    RGB Only & 0.205 & 0.414 &  0.670 & 0.512\\ 
    \hline
    FLS (Ours) & \textbf{0.124} & \textbf{0.457} & \textbf{0.775} &  \textbf{0.575}\\
    \hline
    \end{tabular}
\end{table}


\begin{figure}[!ht]
    \centering
    \setlength{\abovecaptionskip}{-0.12cm}
     \includegraphics[width=1\linewidth]{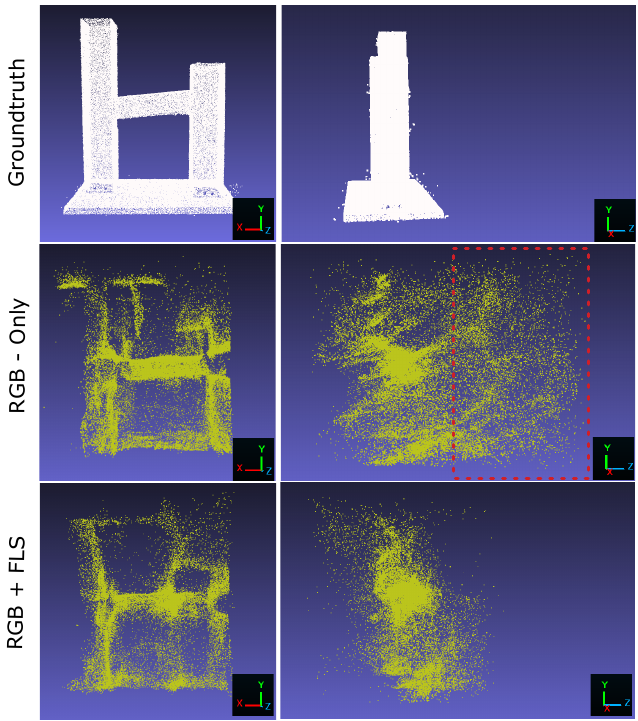}
     \caption{{\bf Experimental reconstructions using RGB-only and RGB+FLS.} When using RGB-only measurements, we observe high error along depth (red box). Integrating FLS measurements improves depth resolution.}
     \label{fig:flsreal}
\end{figure}


%% file: sections/07_conclusion.tex
\section{Conclusion}
\label{sec:conclusion}

In this paper, we introduce a $z$-axis splatting pipeline for Gaussian splatting, which enables the fusion of RGB camera information with various types of depth-resolved acoustic measurements, such as echo-sounders and FLS. 
Through extensive validation on simulated, emulated, and real-world hardware experimental data, we demonstrate superior performance compared to RGB-only methods for both novel view synthesis and geometric reconstruction. By leveraging the depth information provided by acoustic sensors, our method offers a promising approach to reconstructing scenes with small baselines.

Our method currently does not account for scattering. Modeling scattering may produce more accurate reconstructions, particularly in the  underwater setting.
In the fusion step, we used a simple linear model to combine the losses of various imaging modalities. 
Non-linear models and adaptive models that change the relative weights over iterations, such as the ones that Qadri et al.~\cite{qadri2024aoneus} have used, may result in better 3D reconstruction. Exploring various combinations of loss functions could be another interesting future direction.  

While this work focused on $z$-axis splatting for sonar fusion, an interesting future direction is to extend our approach to handle radar and lidar systems, which share a similar acquisition model as sonars. 
Finally, generalizing our method to dynamic scenes and to sensors that operate in this space, such as Doppler cameras and frequency-modulated continuous-wave time-of-flight cameras, could be another interesting extension of our technique.
